\documentclass[12pt]{iopart}
\usepackage{graphicx}
\usepackage{tabularx}
\usepackage{makecell}
\usepackage{array}
\usepackage{hyperref}

\expandafter\let\csname equation*\endcsname\relax

\expandafter\let\csname endequation*\endcsname\relax
\usepackage{amsmath}

\begin{document}

\title[Explainable AI Integrated Feature Engineering for Wildfire Prediction]{Explainable AI Integrated Feature Engineering for Wildfire Prediction }

\author{Di Fan, Ayan Biswas, James Paul Ahrens}

\address{Los Alamos National Laborator, Los Alamos, NM 87545}
\ead{difan@lanl.org, ayan@lanl.gov, ahrens@lanl.gov}
\vspace{10pt}
\begin{indented}
\item[]December 2023
\end{indented}

\begin{abstract}
Wildfires present intricate challenges for prediction, necessitating the use of sophisticated machine learning techniques for effective modeling\cite{jain2020review}. In our research, we conducted a thorough assessment of various machine learning algorithms for both classification and regression tasks relevant to predicting wildfires. We found that for classifying different types or stages of wildfires, the XGBoost model outperformed others in terms of accuracy and robustness. Meanwhile, the Random Forest regression model showed superior results in predicting the extent of wildfire-affected areas, excelling in both prediction error and explained variance. Additionally, we developed a hybrid neural network model that integrates numerical data and image information for simultaneous classification and regression. To gain deeper insights into the decision-making processes of these models and identify key contributing features, we utilized eXplainable Artificial Intelligence (XAI) techniques, including TreeSHAP, LIME, Partial Dependence Plots (PDP), and Gradient-weighted Class Activation Mapping (Grad-CAM). These interpretability tools shed light on the significance and interplay of various features, highlighting the complex factors influencing wildfire predictions. Our study not only demonstrates the effectiveness of specific machine learning models in wildfire-related tasks but also underscores the critical role of model transparency and interpretability in environmental science applications.
\end{abstract}

%
%
%
%
%

\section{Introduction}

Wildfires, while crucial in maintaining the ecological equilibrium of our planet by aiding in carbon capture, promoting biodiversity, and supporting soil and water conservation, face increasing threats from their own intensity and frequency. These natural phenomena, originating in forests, pose significant dangers to the environment, wildlife, and human populations. Forests play an essential role in sustaining life, leading to a worldwide agreement on the urgency of addressing wildfire challenges. The consequences of wildfires extend across environmental, economic, and social realms.

These fires disrupt ecological balance, affecting the flora, fauna, and soil health of the regions they engulf. The lasting impact of wildfires can lead to a decrease in forested areas, which in turn causes soil erosion and changes in plant regrowth patterns. A reduced tree canopy can exacerbate air pollution. The increase in carbon dioxide and monoxide emissions from these fires can contribute to global warming and climate change. Additionally, the smoke from wildfires poses significant health risks, creating new health concerns and exacerbating existing health problems for people.

Economically, the impact of wildfires is profound. Many indigenous and local communities rely heavily on forest resources for their livelihoods. The devastation caused by wildfires can severely affect these groups, compromising their daily subsistence. Additionally, the economic implications extend beyond rural areas, impacting urban sectors as well. The timber industry, for example, faces risks due to the substantial loss of timber caused by fires. In several countries, key trade routes run through forest areas, and any damage due to wildfires can disrupt transportation and trade activities. Furthermore, the tourism sector is also vulnerable; damaged and scarred landscapes from wildfires can dissuade potential tourists, affecting the industry significantly.

In addition to their environmental and economic impacts, wildfires represent a significant risk to human safety. When uncontrolled, these fires can encroach upon nearby residential areas, leading to property damage and potential loss of life. Residents in affected regions may suffer from ongoing health issues. As previously mentioned, the smoke produced by these fires is a health hazard, capable of causing new health problems and exacerbating pre-existing conditions. Particularly at risk are the young and the elderly, who are more susceptible to the adverse effects of wildfire smoke.

In the field of predicting wildfire risks, there's a growing trend of using data-driven machine learning techniques. For example, George \cite{sakr2010artificial} utilized a Support Vector Machine to create an algorithm specifically for fire risk categorization, dividing it into four distinct classes. This algorithm was based on parameters that included historical fire data and specific meteorological conditions. When applied in Lebanon, George's model demonstrated remarkable effectiveness, achieving an impressive 96\% accuracy rate in predicting the hazard levels of fires, using previous weather conditions as a basis for its assessments. Similarly, Onur and colleagues \cite{satir2016mapping} delved into the use of the Multilayer Perceptron, powered by a back-propagation algorithm, to assess wildfire risks. Their research focused on predicting the likelihood of forest fires in Turkey's Upper Seyhan Basin. Complementing this approach, Daniela Stojanova and her team \cite{stojanova2012estimating} compared traditional statistical methods with advanced data mining techniques like decision trees. Their goal was to develop a predictive model for fire occurrences in the Kras region, coastal areas of Slovenia, and beyond. To achieve this, they utilized data from Geographic Information Systems (GIS), Remote Sensing imagery, and sophisticated weather forecasting models .

The field of Explainable Artificial Intelligence (XAI) is rapidly evolving, introducing a wide range of methods and categorizations (e.g., \cite{antoniadi2021current,das2020opportunities,fouladgar2020xai,hu2021model,samek2021explaining,srihari2020explainable}). For instance, Samek et al. \cite{samek2021explaining} categorize XAI algorithms based on various mathematical techniques used for generating explanations, including local surrogates, occlusions, gradient-driven methods, and layer-specific relevance propagation. Furthermore, XAI categorizations are being tailored for specific research areas, such as medical image analysis, as illustrated by Muddamsetty et al. Another approach to XAI classification focuses on its functionality and delivery method, encompassing algorithms, visual representations, auditory methods, Knowledge Graphs (KGs), and straightforward textual explanations, as discussed by Rawal et al \cite{rawal2021recent}.
In the context of wildfire research, two recent studies have explored the application of XAI, specifically targeting the prediction and analysis of wildfire occurrences and their dimensions. These studies represent a significant step in integrating advanced AI methodologies into environmental research, potentially enhancing our understanding and management of wildfire events \cite{al2022machine,cilli2022explainable}.

In conclusion, wildfires significantly impact our environment, economy, and societal health. Prompt detection and intervention are essential to curtail their reach and aftermath. By proactively tackling these fires, we can minimize their adverse effects on our environment, economic frameworks, and societies. This work focuses on utilizing machine learning algorithms for three primary objectives.
\begin{itemize}
    \item \textbf{Prediction of Unintended Burned Land:} We aim to leverage machine learning algorithms to predict the extent of land that might be unintentionally burned inside and outside the intended boundaries of prescribed fires. By analyzing historical fire data, weather patterns, and other relevant factors, the project seeks to provide accurate estimates of potential risks and improve forest management decisions to minimize the impact on nearby communities and ecosystems.
    \item \textbf{Enhanced Prescribed Fire Safety Assessment:} Another key objective is to develop machine learning models for assessing the safety of prescribed fire behavior. By integrating data on fire behavior, weather conditions, fuel types, and terrain characteristics, the project aims to create predictive tools that can identify potential risks and enable fire management teams to optimize burn plans. These safety assessments will contribute to safer and more controlled prescribed burns, reducing hazards to human lives, properties, and the environment.
    \item \textbf{Designing Explainable AI(XAI)  Models:} Additionally, we aim to show machine learning models with strong interpretability and explainability. While machine learning can provide valuable insights, black-box models often lack transparency, hindering forest managers' ability to understand the underlying factors influencing predictions. By focusing on interpretable models, the project seeks to empower decision-makers with a clear understanding of the features and processes driving the model's outputs, facilitating more effective and informed forest fire management strategies.
\end{itemize}
\normalsize

\section{Methods}

In our research, we focus on two primary tasks: Classification and Regression. The Classification task is dedicated to predicting safety behaviors in the context of wildfires, while the Regression task aims at estimating the burned areas both within and outside the fire boundaries. Initially, we tackle these tasks separately, detailing the specific methods and algorithms we choose for each.

For the Classification task, we compare different algorithms and utilize algorithm XGBoost best suited to differentiate and predict various safety behaviors related to wildfires. This involves analyzing patterns and correlations in the data that could indicate effective safety measures.

In the Regression task, our objective is to accurately predict the extent of areas affected by wildfires, both inside and outside the fire's boundaries. This requires a different set of algorithms, Random Forest is the best for processing spatial and environmental data to estimate the impact area of the fire after comparsion.

After conducting these tasks independently, we then employ a Convolutional Neural Network (CNN) model to address both tasks simultaneously. The CNN model, known for its proficiency in handling image data, allows us to integrate and analyze complex datasets. This approach enables us to perform both classification and regression in a unified framework, potentially leading to more cohesive and insightful results in our wildfire research.
\subsection{Classifciation-XGBoost}
XGBoost, abbreviated from ``eXtreme Gradient Boosting", has emerged as a paramount tool in the machine learning landscape, excelling in both classification and regression tasks \cite{chen2016xgboost}.  When juxtaposed with the traditional Gradient Boosting (GB), it emerges superior due to a suite of enhanced design features. These include improved resilience in managing missing values, the incorporation of an approximate and sparsity-conscious split-finding algorithm, capabilities for parallel computing, cache-conscious access patterns, block compression, and sharding. This suite of advancements renders XGBoost an exceedingly effective tool, especially for intricate and precision-critical classification scenarios.

The construction of each decision tree within XGBoost leverages the gradient descent approach. This starts with a predefined threshold and iteratively adjusts the weights to curtail residuals in every iteration. As a result, each tree constructed in a subsequent iteration is distinct, as it aims to rectify or mitigate errors introduced by its predecessor.

\subsection{Regression-Random Forest}
The Random Forest (RF)\cite{breiman2001random} is essentially a collection of multiple decision trees that work collectively as an ensemble. Within this structure, every single tree in the RF offers a class prediction, and the class that amasses the majority of votes across all trees is then selected as the model's final prediction. At the heart of RF lies the principle known as the ``wisdom of crowds." This philosophy suggests that a diverse group of models (in this case, trees) working together can make better collective judgments than any individual model on its own. Essentially, by pooling together multiple ``weak learners" (individual trees), an overarching ``strong learner" can be established.

Much like its predecessor, the Decision Tree (DT) classifier, the RF doesn't necessitate feature scaling. However, it presents distinct advantages over the DT. For instance, RF is more resilient to variations in training samples and any noise within the training dataset. On the flip side, while the DT's decisions can be visualized and understood with relative ease, the complex ensemble nature of RF makes it less intuitive and harder to decipher \cite{belgiu2016random}.

A Random Forest consists of a collection of decision trees, each independently constructed using bootstrapped samples from the training data. During the construction of a tree, at each node, a random subset of features is chosen as candidates for splitting. Final predictions are made by aggregating predictions from individual trees. For classification, majority voting is used, while for regression, predictions are usually averaged.
\subsection{Convolutional Neural Network}

The neural network architecture comprises three main components: ResNet-50, a classification head, and a regression head. Initially, a pre-trained ResNet-50 is loaded, followed by the replacement of its final fully connected layer. Subsequently, both the classification and regression heads are defined. The output from the modified fully connected layer is concatenated with an additional set of five features. These combined features are then fed into both the classification and regression heads, facilitating the computation of respective outputs for classification and regression tasks. This approach leverages the strength of ResNet-50 in feature extraction and expands its applicability through specialized heads for distinct predictive objectives.

Our classification loss $L_{cls}$  and regression loss  $L_{reg}$ are as follows: 
\begin{equation}
L_{cls} = -\frac{1}{N} \sum_{i=1}^{N} [y_i \log(\hat{y}_i) + (1 - y_i) \log(1 - \hat{y}_i)]
\end{equation}

\begin{equation}
L_{reg} = \frac{1}{N} \sum_{i=1}^{N} (y_i - \hat{y}_i)^2
\end{equation}
The total  loss is:
\begin{equation}
    L =  L_{cls} + \lambda L_{reg}
\end{equation}
Here, $y_i$ denotes the true label (either 0 or 1) for the i-th data point, $\hat{y}_i$ is the predicted probability of the 
i-th data point being in class 1. $N$ is the total number of data points in the dataset.
\subsection{Explainable AI methods}

Explainable Artificial Intelligence (XAI) involves the development of AI methods and techniques that produce results interpretable by human experts. In XAI, researchers primarily concentrate on two aspects: Interpretability \cite{arrieta2020explainable} and Explainability \cite{arrieta2020explainable} of AI models. Interpretability is the capacity of a model to present its workings in terms understandable to a human. Ideally, these explanations should be in the form of logical decision rules (if-then statements) or convertible into such rules. The terms used in these explanations should either stem from domain-specific knowledge pertinent to the task or be based on general knowledge relevant to that task. Explainability is a broader and more inclusive concept than interpretability. It refers to a model's ability to clearly articulate its internal mechanics and the rationale behind its decisions, particularly in the case of complex, opaque `black-box' models. Explainability extends beyond interpretability, often serving as a means to achieve a certain degree of interpretability in otherwise intricate models. In summary, explainability seeks to enhance the transparency and comprehensibility of models, thereby making them more accessible and accountable to users and stakeholders.

Here we introduce several XAI methods that we use in our paper.
\subsubsection{TreeSHAP}
TreeSHAP \cite{lundberg2017unified,lundberg2018consistent}is a variant of the SHAP (SHapley Additive exPlanations) framework, designed to clarify machine learning predictions by calculating Shapley values for specific data points, detailing the combined contributions of its distinct feature variables.

In our study, we employed the TreeSHAP method to analyze the predictions made by the Extreme Gradient Boosting algorithm, with the goal of overcoming the typical opacity associated with machine learning algorithms and fostering an Explainable Artificial Intelligence (XAI) framework. By adopting this approach, we aimed not only to clarify the decision-making process of the model but also to pinpoint the key features that significantly impact both the classification of wildfire behavior and the estimation of the burned area. This methodology aids in unraveling the complexities of the algorithm, providing insights into how specific variables influence the model's predictions.

\subsubsection{LIME}
\cite{ribeiro2016should} called Local Interpretable Model-agnostic Explanations, is based on a surrogate model. The surrogate model is usually a linear model constructed based on different samples of the main model. It does this by sampling points around an example and evaluating models at these points. LIME generally computes attribution per sample basis. It takes a sample, perturbs multiple times based on random binary vectors, and computes output scores in the original model. It then uses the binary features (binary vectors) to train an interpretable surrogate model to produce the same outputs. Each of the coefficients in the trained surrogate linear model serves as the input feature's attribution in the input sample.  One of the major issues with LIME is robustness. LIME explanations can disagree if computed multiple times. This disagreement occurs mainly because this interpretation method is estimated with data, causing uncertainty. Moreover, the explanations can be drastically different based on kernel width and feature grouping policies.

\subsubsection{Partial Dependence Plot}
The partial dependence plot\cite{greenwell2017pdp}, often abbreviated as PDP, illustrates the marginal influence that one or two features exert on a machine learning model's predicted result (as noted by J. H. Friedman 200130). This plot can reveal if the association between the target and a specific feature is linear, monotonic, or of a more intricate nature. In the context of a linear regression model, these plots consistently depict a linear relationship. 
\subsection{Grad-CAM}

Gradient-weighted Class Activation Mapping (Grad-CAM) \cite{selvaraju2017grad} creates a coarse localization map that highlights the key areas in the image for concept prediction by using the gradients of each target concept flowing into the final convolutional layer. Here we add Grad-CAM to our classification neural network to show which regions vary significantly during age changes. It determines which area of the feature map has a significant correlation with the classification results by using the output probability to make an inference about it in reverse.
\begin{figure*}[ht!]
\centering
{\includegraphics[width=0.8\textwidth]{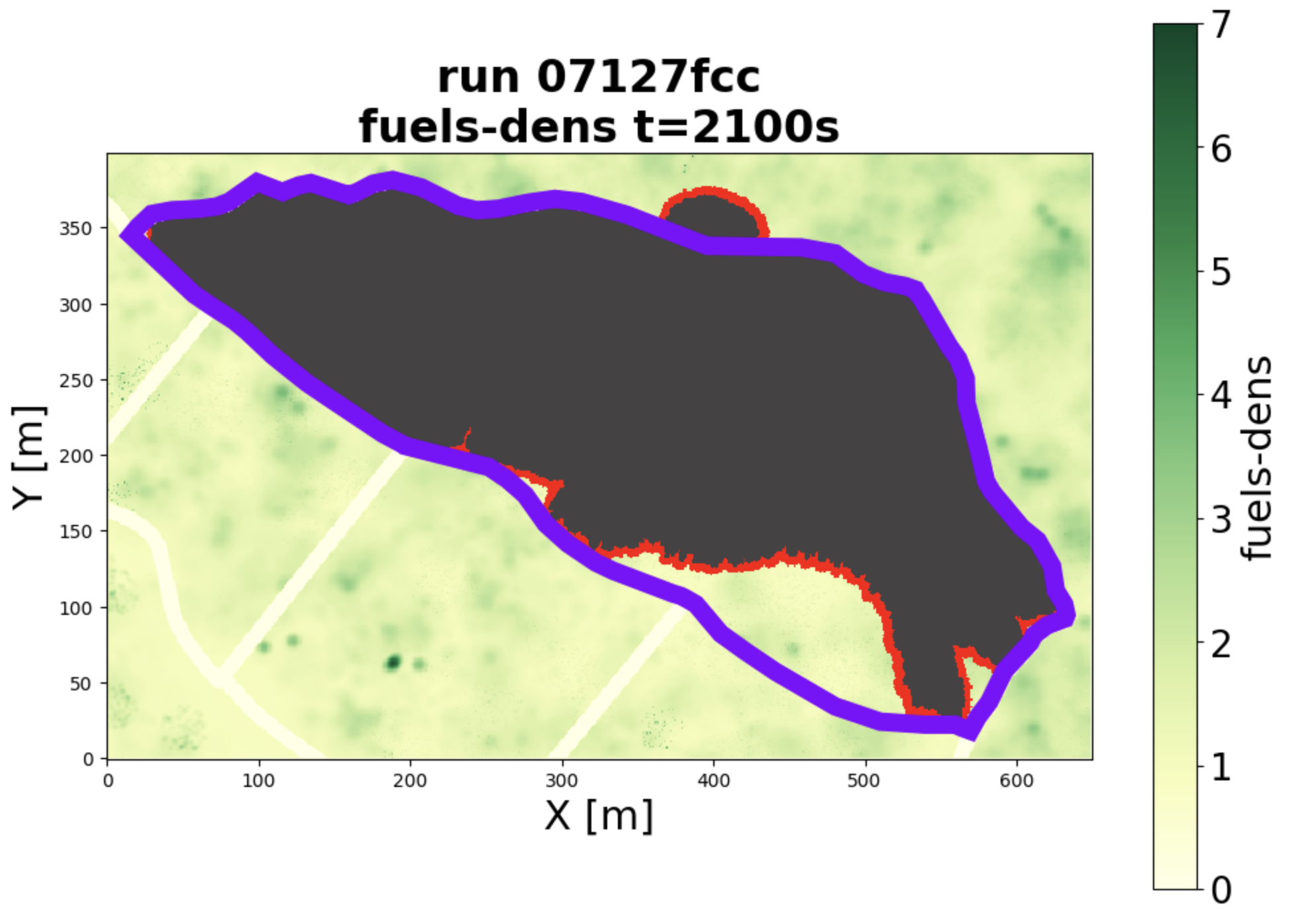}}
\caption{Wildfire simulation images from QUIC-fire.}
\label{fig:wildfire}
\end{figure*}
\begin{figure*}[ht!]
\centering
{\includegraphics[width=0.8\textwidth]{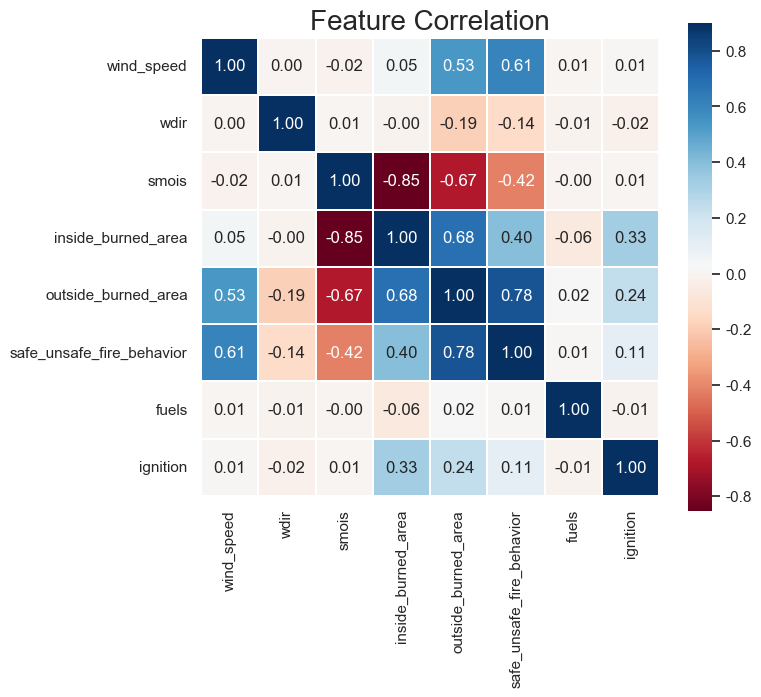}}
\caption{Feature Correlation.}
\label{fig:feature_cor}
\end{figure*}
\section{Experiment}

\subsection{Data}

In our research, we conducted experiments using our proposed methods on a benchmark dataset derived from the public Yosemite wildfire. Our study incorporated five key features: `wind$\_$speed', `wdir' (wind direction), `smois' (soil moisture), `fuels', and `ignition'. These features were selected based on their relevance and potential impact on wildfire behavior and spread.

For the classification task, our primary objective was to predict 
`$safe\_unsafe\_fire\_behavior$'. This involved categorizing fire behavior as either safe or unsafe based on the given conditions. In addition to the traditional data inputs, we also integrated image data into our Convolutional Neural Network (CNN) approach. We utilized visual representations, as depicted in the referenced Figure \ref{fig:wildfire} from QUIC-fire\cite{linn2020quic}, to enhance our predictive capabilities. Regarding the regression task, we aimed to predict two specific outcomes: `$inside\_burned\_area$' and `$outside\_burned\_area$'. These targets represent the quantification of the areas affected by the wildfire, both within and beyond the fire's immediate boundary. 

To better understand the interplay between the selected features and these targets, we refer to Figure \ref{fig:feature_cor}, which illustrates the correlations and relationships between these variables. This visual representation aids in comprehending the intricate dynamics at play in wildfire spread and impact, thereby enriching our analysis and the effectiveness of our predictive models.

\subsection{Evaluation Metrics}
In our study, to evaluate the performance of our machine 
learning models, we considered the analysis of below men
tioned popularly used evaluation metrics.
\begin{itemize}
\item Accuracy: It defines correctly classified areas with susceptible to Landslides. It can be computed as:
\begin{equation}
\text { Accuracy }=\frac{T P+T N}{T P+F P+T N+F N}
\end{equation}
\item Recall: It is defined as the ratio of the number of positive samples that have been correctly predicted as Landslide Susceptible corresponding to all Landslide Susceptible samples in the data. It can computed as:
\begin{equation}
\text { Recall }=\frac{T P}{T P+F N}
\end{equation}
\item Precision: It is defined as the ratio of the number of positive samples that have been correctly predicted as Landslide Susceptible corresponding to all samples predicted as Landslide Susceptible. It can be computed as:
\begin{equation}
\text { Precision }=\frac{T P}{T P+F P}
\end{equation}
where TP,FP, TN, FN are True Positive, False Positive, True Negative and False Negative respectively.
\item F1-Score: It is delineated as the term that balances between recall and precision. It can be defined as;
\begin{equation}
 F 1-\text { Score }=2 \times \frac{\text { Recall } \times \text { Precision }}{\text { Recall }+ \text { Precision }}
\end{equation}
\item Mean Absolute Error (MAE):  It is defined as the mean of the absolute value of the errors:
\begin{equation}
\frac{1}{n} \sum_{i=1}^n\left|y_i-\hat{y}_i\right|
\end{equation}
\item Mean Squared Error (MSE): It is defined as the mean of the squared errors:
\begin{equation}
\frac{\sum_{i=1}^n\left(y_i-\hat{y_i}\right)^2}{N}
\end{equation}

\item Root Mean Squared Error (RMSE): It is defined as the square root of the mean of the squared errors:
\begin{equation}
\sqrt{\frac{\sum_{i=1}^N\left(x_i-\hat{x}_i\right)^2}{N}}
\end{equation}

\item R-squared (R2) score: It is also known as the coefficient of determination, is a statistical measure that represents the proportion of the variance in the dependent variable (target) that is predictable from the independent variables (features) in a regression model. It measures the goodness of fit of the regression model to the data.
\begin{equation}
R^2=1-\frac{R S S}{T S S}
\end{equation}
where $R^2=$coefficient of determination, $R S S=$ sum of squares of residuals, $T S S=$ total sum of squares.

\end{itemize}
\subsection{Parameter Optimization}
Hyperparameter optimization is a critical component in the design of machine learning algorithms, where traditional methods like grid search and random search can often be computationally expensive. Bayesian optimization emerges as a principled strategy for global optimization of expensive black-box functions, making it particularly suitable for hyperparameter tuning. 

Bayesian optimization is an effective method for solving functions that are computationally expensive to find the
extrema[6]. It can be applied for solving a function which does not have a closed-form expression. It can also be
used for functions which are expensive to calculate, the derivatives are hard to evaluate, or the function is nonconvex. Given a machine learning model with hyperparameters $\theta$, and a validation set loss $L(\theta)$, the goal is to find:
\begin{equation}
\theta^*=\arg \min _\theta L(\theta)
\end{equation}
Bayesian optimization assists by constructing a surrogate probabilistic model over the loss function and efficiently searching the hyperparameter space $\Theta$ to minimize the expected validation loss.
Table \ref{tab:1} and \ref{tab:2} shows the parameter space and optimized parameter for classification model respectively.  Table \ref{tab:3} shows the parameter space  for regression model,  Table \ref{tab:4} and \ref{tab:5} shows the optimized parameter for inside burned area and outside burned area model respectively.

To optimize the performance of our Convolutional Neural Network (CNN) model, we engaged in a systematic parameter tuning process. This involved adjusting several key hyperparameters to refine the network's ability to learn from the data effectively and make accurate predictions. The parameters we tuned include:Learning Rate and $\lambda$ for tuning loss.

\begin{table*}
\caption{Search spaces for model hyperparameters for classification.}
\centering
 \begin{tabularx}{\textwidth}{|l|X|}
\hline
\textbf{Model} & \textbf{Parameters} \\
\hline
Logistic Regression & C: $0.1$ to $10$ (log-uniform), Penalty: l1, l2, Solver: liblinear, lbfgs \\
\hline
KNN & n\_neighbors: 3, 5, 7, 9, 11, 13, p: 1, 2 \\
\hline
SVM & C: $0.1$ to $10$ (log-uniform), Kernel: linear, rbf \\
\hline
RandomForest & n\_estimators: 10 to 100, max\_depth: 10 to 20, min\_samples\_split: 2 to 10 \\
\hline
DecisionTree & max\_depth: 10 to 20, min\_samples\_split: 2 to 10 \\
\hline
Xgboost & n\_estimators: 50 to 200, max\_depth: 3 to 5, learning\_rate: $0.01$ to $0.2$ (log-uniform) \\
\hline
Catboost & iterations: 50 to 200, depth: 4 to 8, learning\_rate: $0.01$ to $0.2$ (log-uniform) \\
\hline
\end{tabularx}
\label{tab:1}
\end{table*}

\begin{table*}
\caption{Best parameters for each classification machine learning model.}
\centering
\begin{tabularx}{\textwidth}{|l|X|}
\hline
\textbf{Model} & \textbf{Best Parameters} \\
\hline
Logistic & C: $0.7756$, Penalty: l2, Solver: lbfgs \\
\hline
KNN & n\_neighbors: 7, p: 2 \\
\hline
SVM & C: $4.2149$, Kernel: linear \\
\hline
RandomForest & max\_depth: 14, min\_samples\_split: 8, n\_estimators: 94 \\
\hline
DecisionTree & max\_depth: 14, min\_samples\_split: 8 \\
\hline
Xgboost & learning\_rate: $0.1140$, max\_depth: 3, n\_estimators: 140 \\
\hline
Catboost & depth: 6, iterations: 159, learning\_rate: $0.1636$ \\
\hline
\end{tabularx}
\label{tab:2}
\end{table*}

\begin{table*}
\caption{Search spaces for model hyperparameters for  regression}
    \centering
    \begin{tabularx}{\textwidth}{|l|X|}
    \hline
    \textbf{Model} & \textbf{Parameter Space} \\
    \hline
    ExplainableBoostingRegressor & learning\_rate: Real(0.01, 0.2, prior=`log-uniform') \\
    \hline
    Linear Regression& fit\_intercept: Categorical([True, False]) \\
    \hline
    Logistic Regression& C: Real(0.1, 10, prior=`log-uniform'), penalty: Categorical([`l1', `l2']), solver: Categorical([`liblinear', `lbfgs']) \\
    \hline
    Ridge Regression& alpha: Real(0.01, 10, prior=`log-uniform') \\
    \hline
    Lasso Regression& alpha: Real(0.01, 10, prior=`log-uniform') \\
    \hline
    ElasticNet & alpha: Real(0.01, 10, prior=`log-uniform'), l1\_ratio: Real(0.1, 0.9, prior=`uniform') \\
    \hline
    SVM & C: Real(0.1, 10, prior=`log-uniform'), kernel: Categorical([`linear', `rbf']), gamma: Categorical([`scale', `auto']) \\
    \hline
    RandomForest & n\_estimators: Integer(10, 100), max\_depth: Integer(10, 20), min\_samples\_split: Integer(2, 10) \\
    \hline
    Xgboost & n\_estimators: Integer(50, 200), max\_depth: Integer(3, 5), learning\_rate: Real(0.01, 0.2, prior='`log-uniform') \\
    \hline
    Catboost & iterations: Integer(50, 200), depth: Integer(4, 8), learning\_rate: Real(0.01, 0.2, prior=`log-uniform') \\
    \hline
    \end{tabularx}
    \label{tab:3}
\end{table*}

\begin{table*}
\caption{Best Parameters for Different Models  for Inside Burned Area}
    \centering
    \begin{tabularx}{\textwidth}{|l|X|}
    \hline
     \textbf{Model} & \textbf{Best Params} \\
    \hline
     ExplainableBoostingRegressor & `learning\_rate': 0.114012258603381 \\
    \hline
     Linear Regression& `fit\_intercept': True \\
    \hline
     Logistic Regression& `C': 4.2149456283335, `penalty': `l1', `solver': `liblinear' \\
    \hline
     Ridge Regression& `alpha': 0.16994636371262764 \\
    \hline
     Lasso Regression& `alpha': 3.2521088005944945 \\
    \hline
     ElasticNet & `alpha': 0.2160217783087772, `l1\_ratio': 0.8349780173355017 \\
    \hline
     SVM & `C': 4.7290805470559185, `gamma': `scale', `kernel': `linear' \\
    \hline
     RandomForest & `max\_depth': 18, `min\_samples\_split': 3, 'n\_estimators': 64 \\
    \hline
     Xgboost & `learning\_rate': 0.12287608582119026, `max\_depth': 5, `n\_estimators': 96 \\
    \hline
     Catboost & `depth': 6, `iterations': 159, `learning\_rate': 0.16356457461011642 \\
    \hline
    \end{tabularx}
    \label{tab:4}
\end{table*}

\begin{table*}
 \caption{Best Parameters for Different Models for Outside Burned Area}
    \centering
    \begin{tabularx}{\textwidth}{|l|X|}
    \hline
     \textbf{Model} & \textbf{Best Params} \\
    \hline
     ExplainableBoostingRegressor & `learning\_rate': 0.03790883530488498 \\
    \hline
     Linear & `fit\_intercept': True \\
    \hline
     Logistic & `C': 4.2149456283335, `penalty':`l1', `solver': `liblinear' \\
    \hline
     Ridge & 'alpha': 0.16994636371262764 \\
    \hline
     Lasso & 'alpha': 2.5041499136197736 \\
    \hline
     ElasticNet & `alpha': 0.2160217783087772, `l1\_ratio': 0.8349780173355017 \\
    \hline
     SVM & `C': 4.7290805470559185, `gamma': `scale', `kernel': `linear' \\
    \hline
     RandomForest & `max\_depth': 18, `min\_samples\_split': 3, `n\_estimators': 64 \\
    \hline
     Xgboost & `learning\_rate': 0.12287608582119026, `max\_depth': 5, `n\_estimators': 96 \\
    \hline
     Catboost & `depth': 6, `iterations': 159, `learning\_rate': 0.16356457461011642 \\
    \hline
    \end{tabularx}
  \label{tab:5} 
\end{table*}

\section{Results and Discussion}

\subsection{Classification}
\begin{figure*}[ht!]
\centering
{\includegraphics[width=0.8\textwidth]{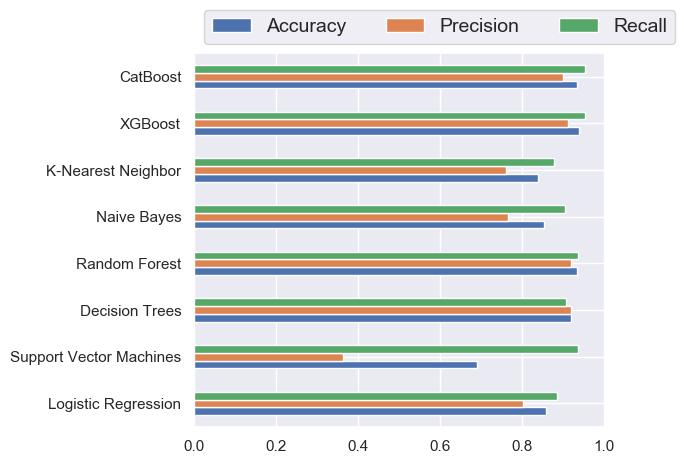}}
\caption{Performance evaluation with all features with best hyperparameters of classifiers}
\label{fig:P}
\end{figure*}
Precision, Recall, F1-Score and Accuracy scores of the 
machine learning classifers is presented on the Figure \ref{fig:P}. We could see the XGBoost achieves the best result, so in this case we used XGBoost as the best model and analyze the results.

One thing needs to be pointed out is that ANN does not do well for the classification with accuracy 0.833, precision 0.90 and recall 0.70. The neural network model is a typical black-box model. Through internal non-linear transformations, the model obscures the correlation between inputs and outputs. Currently, the interpretability of neural networks is a hot research topic, but there is no universally recognized mainstream method to interpret the parameters and outputs of neural networks. Neural networks may not perform optimally due to the data being either linearly separable or very close to it. The predominant factors in the predictions seem to revolve around surface moisture and wind speed, making the data appear ``too straightforward" for a neural network to effectively learn from.
\begin{figure}
\centering
{\includegraphics[width=0.4\textwidth]{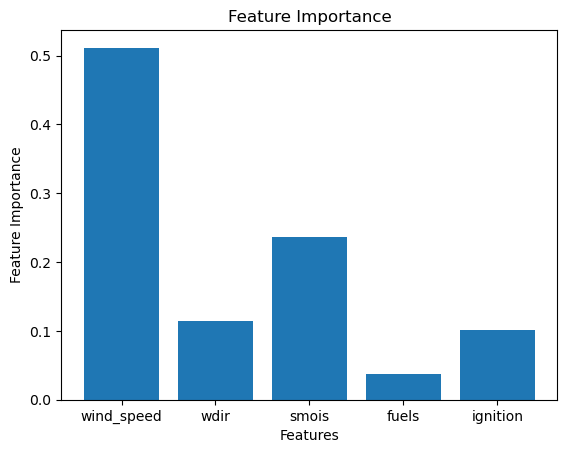}}
\caption{Feature Importance for XGBoost classifier.}
\label{fig:fi}
\end{figure}

The XGBoost method learns in a recursive manner using decision tree models. At each level, it uses a decision tree model to learn the residuals that the previous model could not handle correctly. In each decision tree, the closer a feature is to the root node, the greater its importance. Therefore, by averaging the importance of each feature in these decision trees, we can obtain the importance of each feature in the entire model. However, this importance reflects the ranking relationship of the feature's impact on the target, rather than the quantitative relationship between the feature and the target. We could see the $wind\_speed$ plays an important role for the feature importance in Figure \ref{fig:fi}.
\begin{figure}
\centering
{\includegraphics[width=0.5\textwidth]{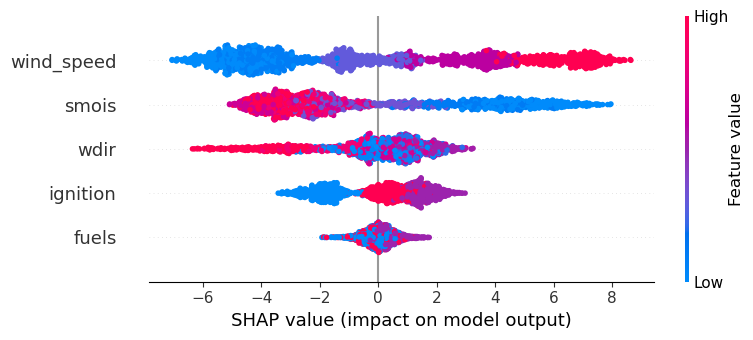}}
\caption{SHAP value(impact on model output)}
\label{fig:shap_c}
\end{figure}
\begin{figure}
\centering
{\includegraphics[width=0.5\textwidth]{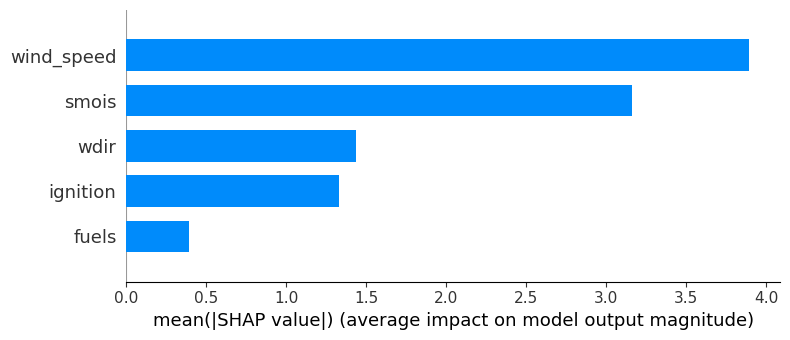}}
\caption{Average impact on model output magnitude}
\label{fig:aver_c}
\end{figure}

The Shapley analysis method can analyze the impact of a specific feature value on the target in a black-box model. Specifically, it calculates the difference between the target value computed by the black-box model when the feature takes the value, and the target value computed by the black-box model when the feature takes its mathematical expectation. This is a quantitative analysis tool. For example, from the above Figure \ref{fig:shap_c}, \ref{fig:aver_c}, we can see that when $wind\_speed$ takes the minimum value, the target value calculated by the black box is 6 units lower than the average. When $wind\_speed$ takes the maximum value, the target value calculated by the black box is about 8 units higher than the average. The core idea of the Shapley method is the quantitative impact on the target value when changing the value of the feature while other features are at their mathematical expectation levels. From this analysis method, in this case, $wind\_speed$ has the greatest impact on the target and they have a positive relationship. $smois$ has a slightly lower impact on the target and they have a negative relationship. When $ignition$ takes a lower value, the target value is relatively small, but the relationship between it and the target value is unclear when it takes a medium or larger value. The other two features have little relationship with the target.

\begin{figure}
\centering
{\includegraphics[width=0.8\textwidth]{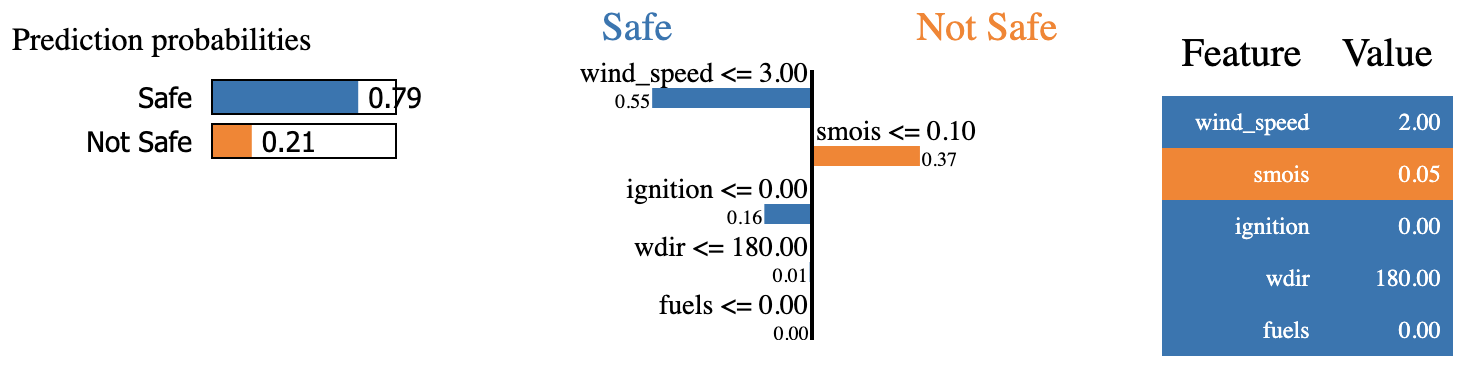}}
\caption{LIME for classification case.}
\label{fig:lime}
\end{figure}
The LIME (Local Interpretable Model-agnostic Explanations) algorithm uses local linear approximation to analyze the local interpretability of a black-box model. For example, here we show one sample in Figure \ref{fig:lime} and analyzed the local linearity of the XGBoost model around them. In the first example, a linear model is used to approximate the prediction results given by the XGBoost model near this sample point. The weight of $wind\_speed$ is -2, $smois$ is 0.05, $wdir$ is -180, and the weights of the other two indicators are 0. The core difference between the LIME algorithm and the Shapley algorithm is that the LIME algorithm explains the local characteristics of the black-box model, while the Shapley algorithm explains the average characteristics of the black-box model.

\begin{figure*}
\centering
{\includegraphics[width=1.0\textwidth]{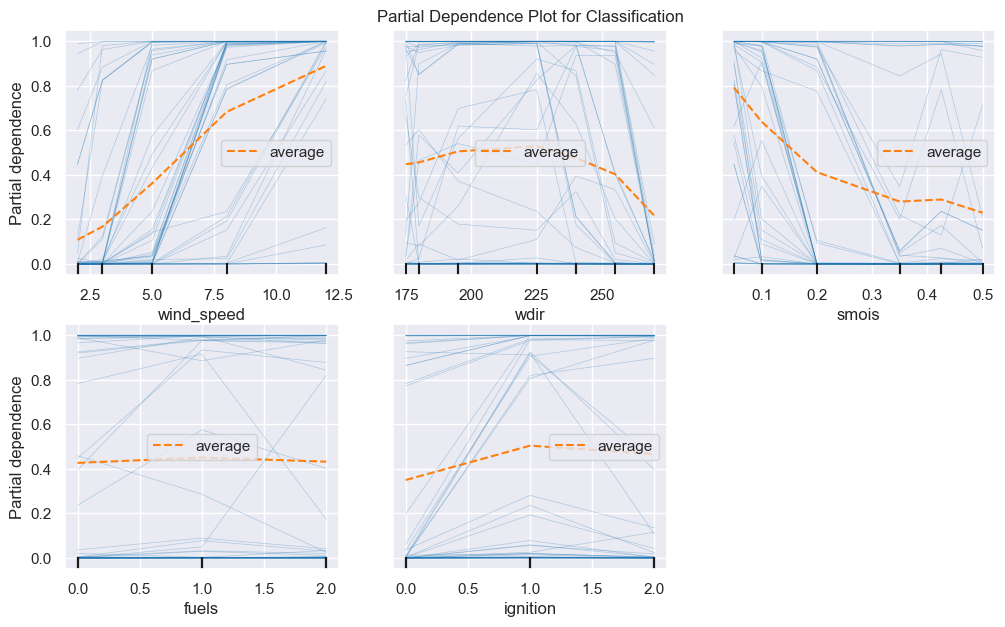}}
\caption{Partial Dependence Plot (PDP) for classification.}
\label{fig:pc}
\end{figure*}
A Partial Dependence Plot (PDP) provides a graphical depiction of the marginal effect of a feature on the predicted outcome of a machine learning model. The PDP is computed after the model has been trained and can offer insights into the relationship between a feature and the predicted outcome. We could see the `wind\_speed' has a positive relationship for output and `smois' has negative relationship for output in Figure \ref{fig:pc}.

\subsection{Regression }
\begin{figure}
\centering
{\includegraphics[width=0.5\textwidth]{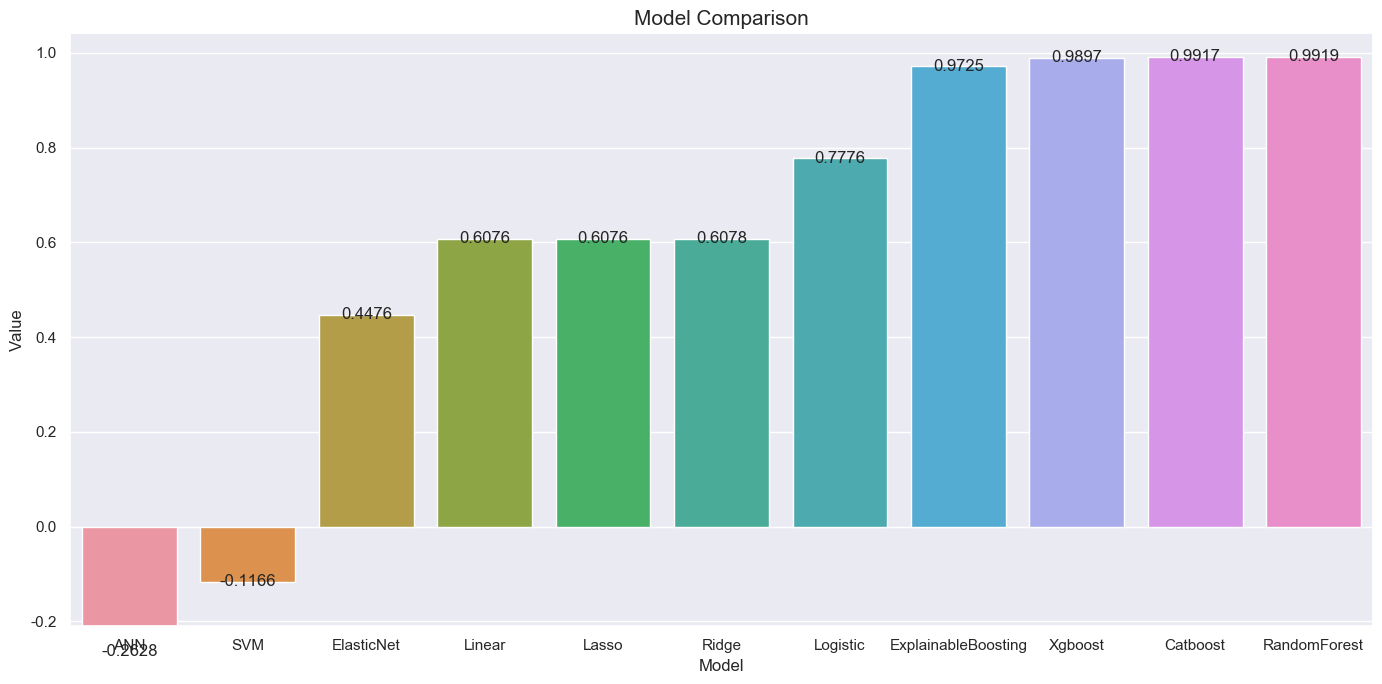}}
\caption{R2 for different inside regression models performance.}
\label{fig:r2_in}
\end{figure}

\begin{figure}
\centering
{\includegraphics[width=0.5\textwidth]{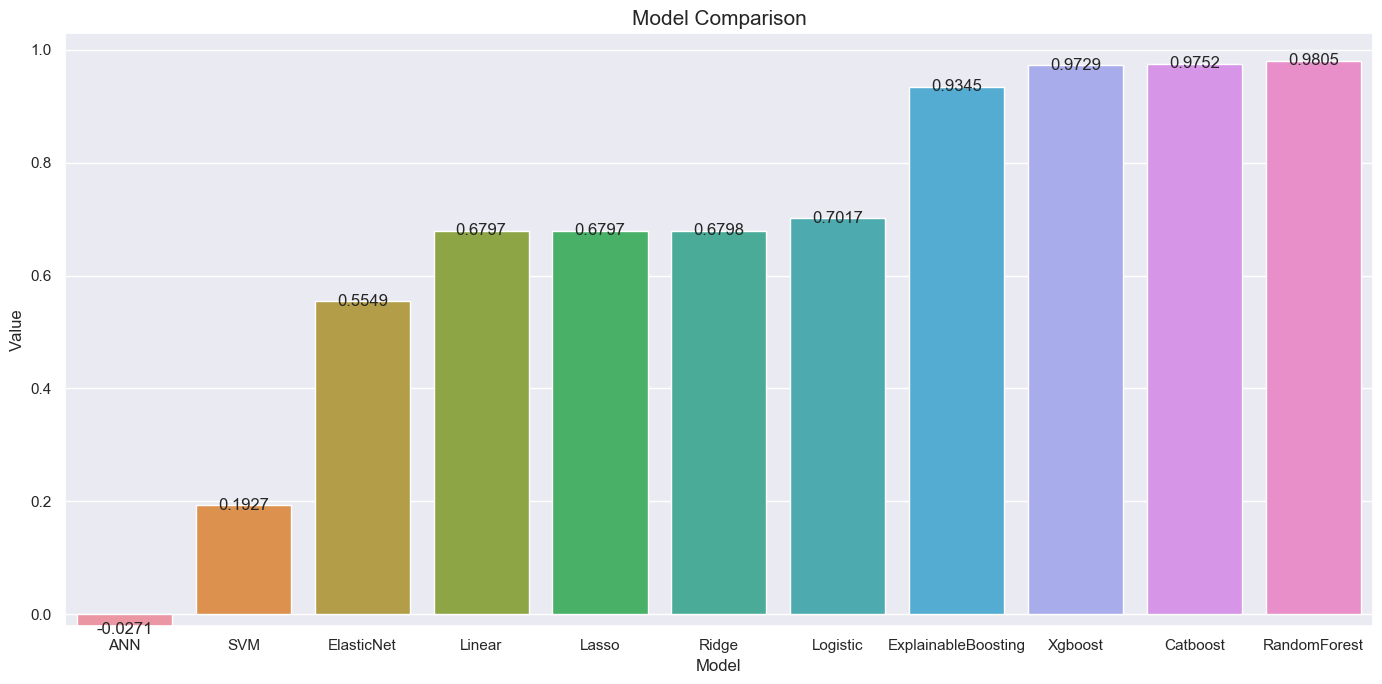}}
\caption{R2 for different outside regression models performance.}
\label{fig:r2_out}
\end{figure}

\begin{figure}
\centering
{\includegraphics[width=0.5\textwidth]{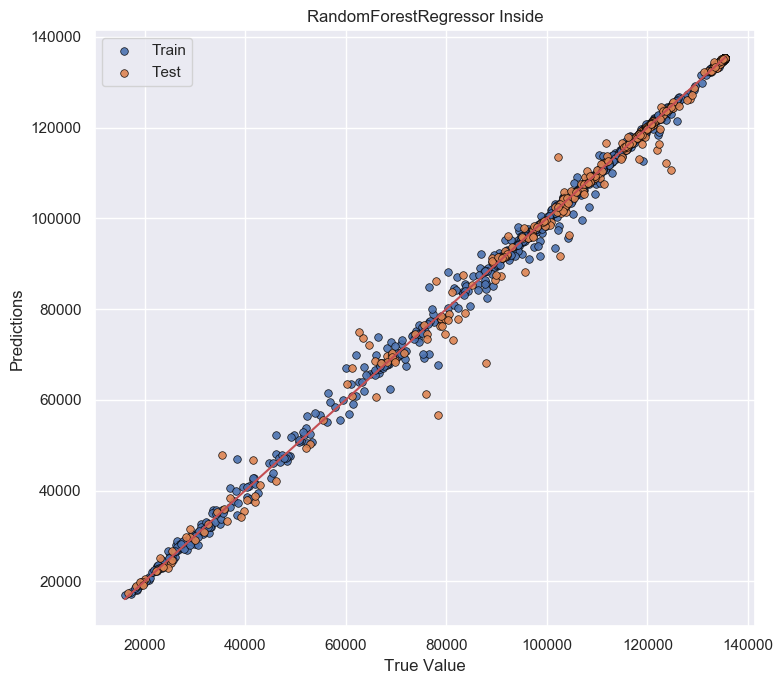}}
\caption{Regression results for inside burned area.}
\label{fig:fit_in}
\end{figure}

\begin{figure}
\centering
{\includegraphics[width=0.5\textwidth]{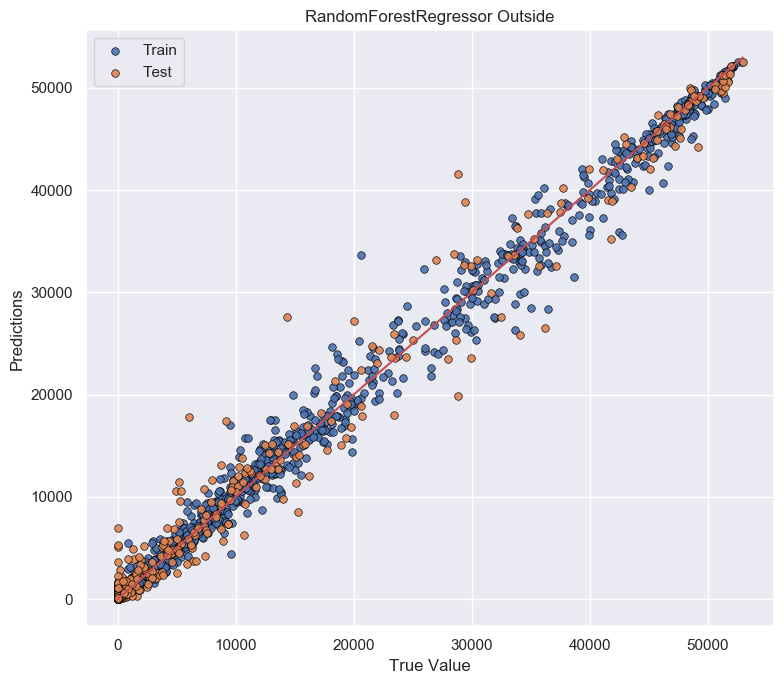}}
\caption{Regression results for outside burned area.}
\label{fig:fit_out}
\end{figure}

R2 scores of the
machine learning regression is presented on the Figure \ref{fig:r2_in} and \ref{fig:r2_out}. We could see the Random Forest Regression achieves the best results for both inside and outside cases. We plot the regression results on Figure \ref{fig:fit_in} and Figure \ref{fig:fit_out} and we could see the fitting results are ideal.
\begin{figure}
\centering
{\includegraphics[width=0.5\textwidth]{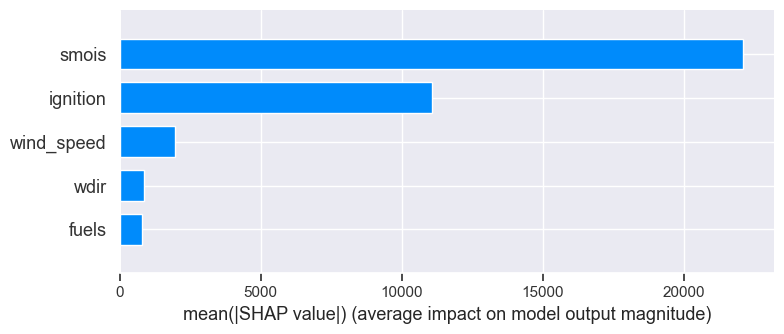}}
\caption{Average impact on model output magnitude for inside burned area.}
\label{fig:fit_ave_in}
\end{figure}
\begin{figure}
\centering
{\includegraphics[width=0.5\textwidth]{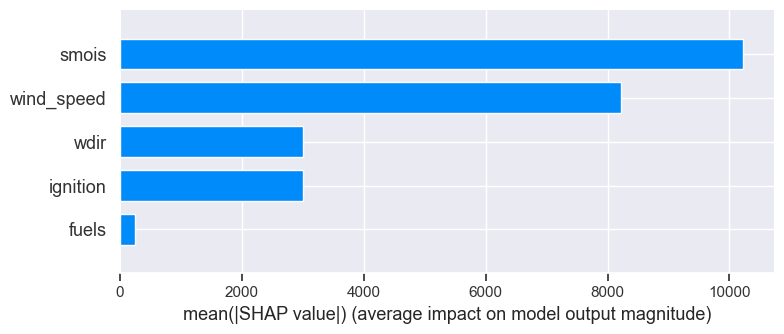}}
\caption{Average impact on model output magnitude for outside burned area.}
\label{fig:fit_ave_out}
\end{figure}

Here, TreeSHAP analysis based on Random Forest Regression has been used to explain the predictions made by the model. It provides insights into the contribution of each feature in influencing individual predictions. By analyzing SHAP values, one can understand the importance and impact of different features on the model's decision-making process, gaining valuable insights into the relationships between features and the target variable in the context of Random Forest Regression. From this analysis method, in the inside case, $smois$ and $ignition$ have the greatest impact on the target and they have a positive relationship. $wdir$ and $fuels$ do not have enough impact on model output. For outside fire area prediction, the $smois$ and $wind\_speed$ are more important in Figure \ref{fig:fit_ave_in} and \ref{fig:fit_ave_out}.
\begin{figure}
\centering
{\includegraphics[width=0.8\textwidth]{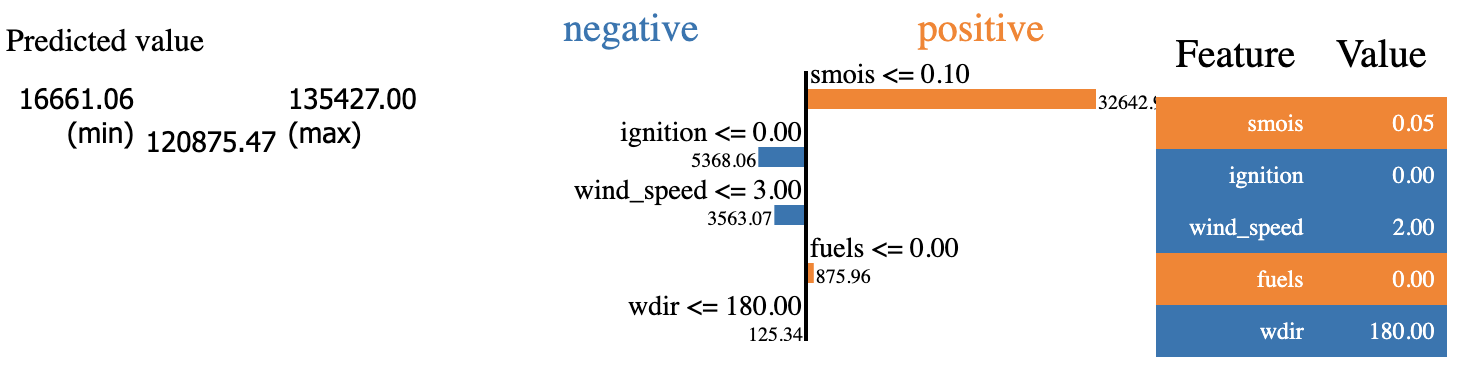}}
\caption{LIME for input burned area regression case.}
\label{fig:lime_in}
\end{figure}
\begin{figure}
\centering
{\includegraphics[width=0.8\textwidth]{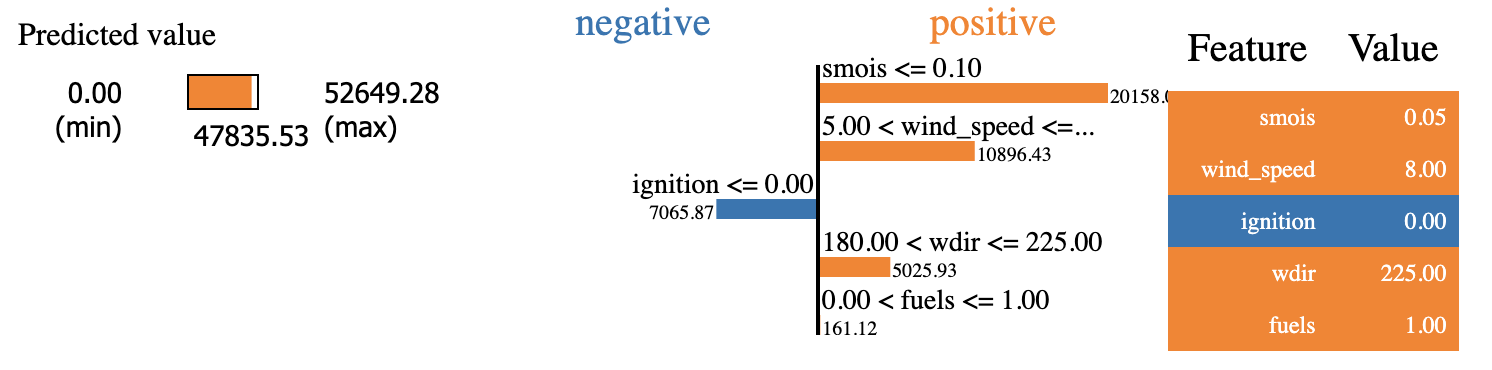}}
\caption{LIME for output burned area regression case.}
\label{fig:lime_out}
\end{figure}
Here we show one example for inside and outside respectively in Figure \ref{fig:lime_in} and Figure \ref{fig:lime_out}. and analyzed the local linearity of the Random Forest Regression model around them. To sum up, for inside fire area, the $smois$ and $ignition$ influence most, for outside fire area, $smois$ and $wind\_speed$ influence output more.

\begin{figure*}
\centering
{\includegraphics[width=1\textwidth]{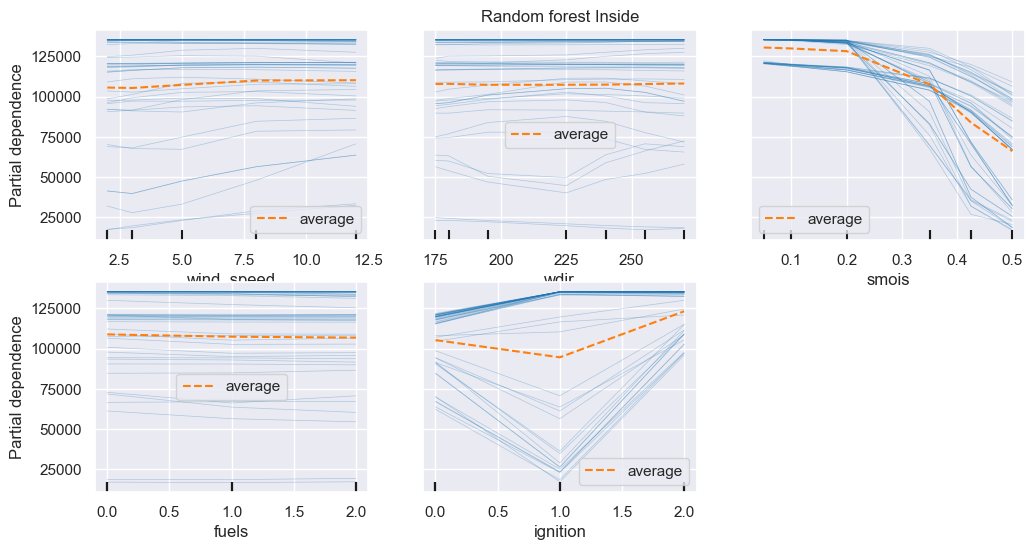}}
\caption{PDP for inside burned area regression case.}
\label{fig:pdp_in}
\end{figure*}
\begin{figure*}
\centering
{\includegraphics[width=1\textwidth]{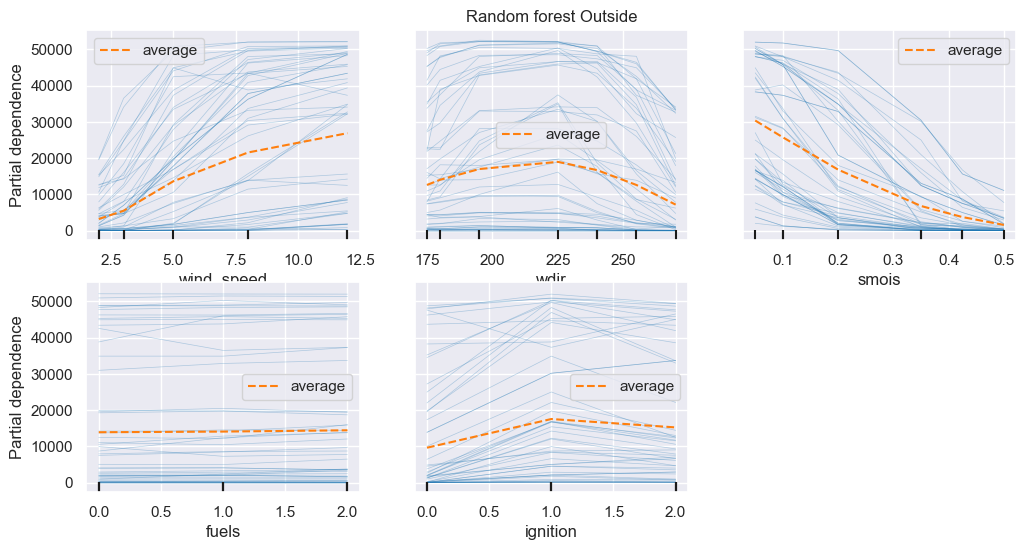}}
\caption{PDP for outside burned area regression case.}
\label{fig:pdp_out}
\end{figure*}

In the PDP Figure \ref{fig:pdp_in} and \ref{fig:pdp_out}, to analyze whether a feature's impact goes up or down, look at the direction of the slope of the PDP. If the slope is positive, it means that as the feature's value increases, the predicted target variable generally goes up. Conversely, if the slope is negative, the predicted target variable goes down as the feature's value increases.

We could see when $smois$ is in a small value region, it has biggest influence for both inside and outside fire area prediction. The $wind\_speed$ will influence the outside fire area when it has a larger value (over 10).

\subsection{CNN}
\begin{figure}
\centering
{\includegraphics[width=0.9\textwidth]{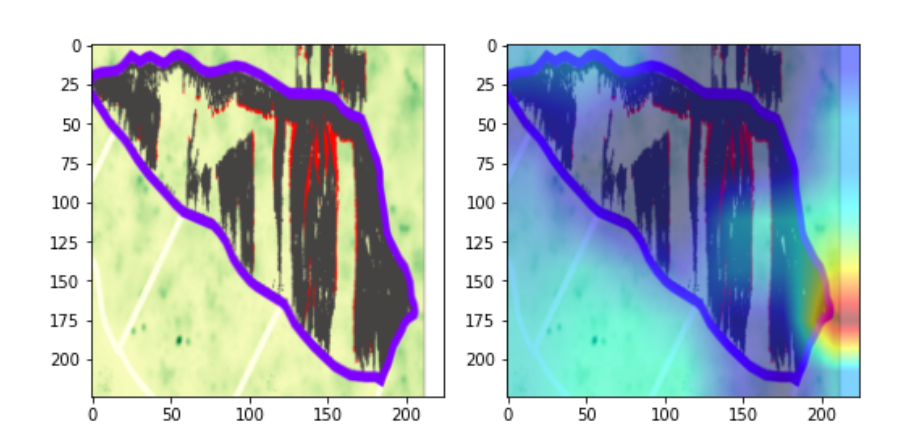}}
\caption{Grad-CAM for prediction.}
\label{fig:cam}
\end{figure}

After tuning parameters, we found learning rate 0.01, $\lambda$ is 5 will achieve the better results. The CNN classification result is 93.10\%, and regression error results for inside and outside burned area are 0.75\% and 2.41\%. Since we utilize the images for classification, we leveraged the Grad-CAM technique to elucidate the classification decisions of our model. This approach allowed us to visually interpret which regions of the input image were most influential in the model's classification output, thereby providing an intuitive understanding of the model's focus and rationale in distinguishing different classes. Grad-CAM's ability to highlight salient features directly on the input images significantly contributes to the transparency and interpretability of our model's decision-making process in classifying complex data sets. From Figure \ref{fig:cam}, we could see the last layer output from the neural network, the red part in the right figure will help us distinguish the inside and outside.

\section{Conclusion}
Our study emphasizes the crucial role of advanced machine learning techniques in addressing the complex challenges of wildfire prediction. We observed the XGBoost model's exemplary performance in classification tasks and the Random Forest model's effectiveness in regression tasks, highlighting the necessity for models specifically tailored to the unique demands of each predictive task. Furthermore, we integrated a Convolutional Neural Network (CNN) capable of concurrently executing classification and regression tasks, achieving results comparable to those of specialized models. The inclusion of eXplainable Artificial Intelligence (XAI) methods further enriched our understanding of these models, uncovering key features and their interactions that influence predictions. As wildfires continue to pose significant environmental and societal threats, the insights from this research underscore the importance of not only precise predictive modeling but also the critical need for interpretability and adaptability in models, ensuring their trustworthiness and effective application in real-world scenarios.
\section{Data availability statement}
The data that support the findings of this study are
from \url{wifire-data.sdsc.edu} and \url{https://burnpro3d.sdsc.edu/index.html}.
\section{Acknowledgments}
We would like to thank Dr. Rodman Linn and the wildfire team for their support in this study.
\section{Reference}
{\small
\bibliographystyle{IEEEtran}
\bibliography{egbib}

\begin{thebibliography}{10}
\providecommand{\url}[1]{#1}
\csname url@samestyle\endcsname
\providecommand{\newblock}{\relax}
\providecommand{\bibinfo}[2]{#2}
\providecommand{\BIBentrySTDinterwordspacing}{\spaceskip=0pt\relax}
\providecommand{\BIBentryALTinterwordstretchfactor}{4}
\providecommand{\BIBentryALTinterwordspacing}{\spaceskip=\fontdimen2\font plus
\BIBentryALTinterwordstretchfactor\fontdimen3\font minus \fontdimen4\font\relax}
\providecommand{\BIBforeignlanguage}[2]{{%
\expandafter\ifx\csname l@#1\endcsname\relax
\typeout{** WARNING: IEEEtran.bst: No hyphenation pattern has been}%
\typeout{** loaded for the language `#1'. Using the pattern for}%
\typeout{** the default language instead.}%
\else
\language=\csname l@#1\endcsname
\fi
#2}}
\providecommand{\BIBdecl}{\relax}
\BIBdecl

\bibitem{jain2020review}
P.~Jain, S.~C. Coogan, S.~G. Subramanian, M.~Crowley, S.~Taylor, and M.~D. Flannigan, ``A review of machine learning applications in wildfire science and management,'' \emph{Environmental Reviews}, vol.~28, no.~4, pp. 478--505, 2020.

\bibitem{sakr2010artificial}
G.~E. Sakr, I.~H. Elhajj, G.~Mitri, and U.~C. Wejinya, ``Artificial intelligence for forest fire prediction,'' in \emph{2010 IEEE/ASME international conference on advanced intelligent mechatronics}.\hskip 1em plus 0.5em minus 0.4em\relax IEEE, 2010, pp. 1311--1316.

\bibitem{satir2016mapping}
O.~Satir, S.~Berberoglu, and C.~Donmez, ``Mapping regional forest fire probability using artificial neural network model in a mediterranean forest ecosystem,'' \emph{Geomatics, Natural Hazards and Risk}, vol.~7, no.~5, pp. 1645--1658, 2016.

\bibitem{stojanova2012estimating}
D.~Stojanova, A.~Kobler, P.~Ogrinc, B.~{\v{Z}}enko, and S.~D{\v{z}}eroski, ``Estimating the risk of fire outbreaks in the natural environment,'' \emph{Data mining and knowledge discovery}, vol.~24, pp. 411--442, 2012.

\bibitem{antoniadi2021current}
A.~M. Antoniadi, Y.~Du, Y.~Guendouz, L.~Wei, C.~Mazo, B.~A. Becker, and C.~Mooney, ``Current challenges and future opportunities for xai in machine learning-based clinical decision support systems: a systematic review,'' \emph{Applied Sciences}, vol.~11, no.~11, p. 5088, 2021.

\bibitem{das2020opportunities}
A.~Das and P.~Rad, ``Opportunities and challenges in explainable artificial intelligence (xai): A survey,'' \emph{arXiv preprint arXiv:2006.11371}, 2020.

\bibitem{fouladgar2020xai}
N.~Fouladgar and K.~Fr{\"a}mling, ``Xai-pt: a brief review of explainable artificial intelligence from practice to theory,'' \emph{arXiv preprint arXiv:2012.09636}, 2020.

\bibitem{hu2021model}
X.~Hu, L.~Chu, J.~Pei, W.~Liu, and J.~Bian, ``Model complexity of deep learning: A survey,'' \emph{Knowledge and Information Systems}, vol.~63, pp. 2585--2619, 2021.

\bibitem{samek2021explaining}
W.~Samek, G.~Montavon, S.~Lapuschkin, C.~J. Anders, and K.-R. M{\"u}ller, ``Explaining deep neural networks and beyond: A review of methods and applications,'' \emph{Proceedings of the IEEE}, vol. 109, no.~3, pp. 247--278, 2021.

\bibitem{srihari2020explainable}
S.~N. Srihari, ``Explainable artificial intelligence,'' \emph{Journal of the Washington Academy of Sciences}, vol. 106, no.~4, pp. 9--38, 2020.

\bibitem{rawal2021recent}
A.~Rawal, J.~McCoy, D.~B. Rawat, B.~M. Sadler, and R.~S. Amant, ``Recent advances in trustworthy explainable artificial intelligence: Status, challenges, and perspectives,'' \emph{IEEE Transactions on Artificial Intelligence}, vol.~3, no.~6, pp. 852--866, 2021.

\bibitem{al2022machine}
M.~K. Al-Bashiti and M.~Naser, ``Machine learning for wildfire classification: Exploring blackbox, explainable, symbolic, and smote methods,'' \emph{Natural Hazards Research}, vol.~2, no.~3, pp. 154--165, 2022.

\bibitem{cilli2022explainable}
R.~Cilli, M.~Elia, M.~D’Este, V.~Giannico, N.~Amoroso, A.~Lombardi, E.~Pantaleo, A.~Monaco, G.~Sanesi, S.~Tangaro \emph{et~al.}, ``Explainable artificial intelligence (xai) detects wildfire occurrence in the mediterranean countries of southern europe,'' \emph{Scientific reports}, vol.~12, no.~1, p. 16349, 2022.

\bibitem{chen2016xgboost}
T.~Chen and C.~Guestrin, ``Xgboost: A scalable tree boosting system,'' in \emph{Proceedings of the 22nd acm sigkdd international conference on knowledge discovery and data mining}, 2016, pp. 785--794.

\bibitem{breiman2001random}
L.~Breiman, ``Random forests,'' \emph{Machine learning}, vol.~45, pp. 5--32, 2001.

\bibitem{belgiu2016random}
M.~Belgiu and L.~Dr{\u{a}}gu{\c{t}}, ``Random forest in remote sensing: A review of applications and future directions,'' \emph{ISPRS journal of photogrammetry and remote sensing}, vol. 114, pp. 24--31, 2016.

\bibitem{arrieta2020explainable}
A.~B. Arrieta, N.~D{\'\i}az-Rodr{\'\i}guez, J.~Del~Ser, A.~Bennetot, S.~Tabik, A.~Barbado, S.~Garc{\'\i}a, S.~Gil-L{\'o}pez, D.~Molina, R.~Benjamins \emph{et~al.}, ``Explainable artificial intelligence (xai): Concepts, taxonomies, opportunities and challenges toward responsible ai,'' \emph{Information fusion}, vol.~58, pp. 82--115, 2020.

\bibitem{lundberg2017unified}
S.~M. Lundberg and S.-I. Lee, ``A unified approach to interpreting model predictions,'' \emph{Advances in neural information processing systems}, vol.~30, 2017.

\bibitem{lundberg2018consistent}
S.~M. Lundberg, G.~G. Erion, and S.-I. Lee, ``Consistent individualized feature attribution for tree ensembles,'' \emph{arXiv preprint arXiv:1802.03888}, 2018.

\bibitem{ribeiro2016should}
M.~T. Ribeiro, S.~Singh, and C.~Guestrin, ``" why should i trust you?" explaining the predictions of any classifier,'' in \emph{Proceedings of the 22nd ACM SIGKDD international conference on knowledge discovery and data mining}, 2016, pp. 1135--1144.

\bibitem{greenwell2017pdp}
B.~M. Greenwell \emph{et~al.}, ``pdp: An r package for constructing partial dependence plots.'' \emph{R J.}, vol.~9, no.~1, p. 421, 2017.

\bibitem{selvaraju2017grad}
R.~R. Selvaraju, M.~Cogswell, A.~Das, R.~Vedantam, D.~Parikh, and D.~Batra, ``Grad-cam: Visual explanations from deep networks via gradient-based localization,'' in \emph{Proceedings of the IEEE international conference on computer vision}, 2017, pp. 618--626.

\bibitem{linn2020quic}
R.~R. Linn, S.~L. Goodrick, S.~Brambilla, M.~J. Brown, R.~S. Middleton, J.~J. O'Brien, and J.~K. Hiers, ``Quic-fire: A fast-running simulation tool for prescribed fire planning,'' \emph{Environmental Modelling \& Software}, vol. 125, p. 104616, 2020.

\end{thebibliography}
}

\end{document}